\documentclass[runningheads]{llncs}
\usepackage{graphicx}
\usepackage{microtype}
\usepackage{wrapfig}
\usepackage{subcaption}
\usepackage{booktabs}
\usepackage{hyperref}
\usepackage{amsmath}
\usepackage{amssymb}
\usepackage[bottom]{footmisc}
\usepackage{mathtools}
\usepackage[capitalize,noabbrev]{cleveref}
\DeclareMathOperator{\EX}{\mathbb{E}}
\usepackage[textsize=tiny]{todonotes}
\newcommand{\repeatthanks}{\textsuperscript{\thefootnote}}
\begin{document}
\title{Learning Generative Factors of EEG Data with Variational auto-encoders}
\titlerunning{Learning Generative Factors of EEG Data with VAEs}
%
%
\author{Maksim Zhdanov\inst{1,2} \and
Saskia Steinmann\inst{3} \thanks{Equal contribution.}  \and
Nico Hoffmann\inst{1} \repeatthanks}
\authorrunning{Zhdanov et al.}
%
\institute{Helmholtz-Zentrum Dresden-Rossendorf, Dresden, Germany \and
TU Dresden, Dresden, Germany \and
Psychiatry Neuroimaging Branch, Department of Psychiatry and Psychotherapy, University Medical Center Hamburg-Eppendorf, Hamburg, Germany}
\maketitle              
\begin{abstract}
Electroencephalography produces high-dimensional, stochastic data from which it might be challenging to extract high-level knowledge about the phenomena of interest. We address this challenge by applying the framework of variational auto-encoders to 1) classify multiple pathologies and 2) recover the neurological mechanisms of those pathologies in a data-driven manner. Our framework learns generative factors of data related to pathologies. We provide an algorithm to decode those factors further and discover how different pathologies affect observed data. We illustrate the applicability of the proposed approach to identifying schizophrenia, either followed or not by auditory verbal hallucinations. We further demonstrate the ability of the framework to learn disease-related mechanisms consistent with current domain knowledge. We also compare the proposed framework with several benchmark approaches and indicate its classification performance and interpretability advantages.

\end{abstract}
\section{Introduction}
\label{introduction}

Analysis of neurological processes in the human brain is a challenging process addressed by neuroimaging. Here, one typically obtains high-dimensional stochastic data, which encourages the usage of machine learning algorithms. In recent years, deep learning discriminative models have been actively applied to neuroimaging issues (see \cite{Davatzikos2019MachineLI} for a review). They yielded state-of-the-art results in classification problems on a variety of benchmark datasets \cite{Zhang2020ASO}, \cite{Wahlang2022BrainMR}, \cite{Li2021ANB}. One downside of deep learning-based classifiers is that they operate as black boxes \cite{Quinn2022TheTG} meaning that interpreting their predictions is often severely complicated.

\cite{Davatzikos2019MachineLI} suggested that hybrid generative-discriminative models might help resolve the issue. Such models can learn low-dimensional representations of data where each dimension corresponds to an independent \emph{generative factor} (i.e. a disentangled representation, see \cite{Liu2021LearningDR} for a review). The discriminative part of the model then forces those factors to capture label information from data \cite{Joy2021CapturingLC}. \emph{Interpretability} is thus achieved via decoding the meaning of generative factors related to particular labels \cite{Higgins2017betaVAELB}. It is especially relevant in neuroimaging as one can observe how underlying pathologies govern the process of data generation.

The paper follows the intuition regarding hybrid generative-discriminative models for neuroimaging data, with particular application to EEG data. Our main contributions are as follows:
\begin{enumerate}
    \item We demonstrate how one can apply characteristic capturing variational auto-encoders (CCVAEs) \cite{Joy2021CapturingLC} to the interpretable classification of EEG data;
    \item We compare the model to two generative models previously used for EEG data: conditional VAEs and VAEs with downstream classification;
    \item We propose an algorithm for decoding generative factors learned by CCVAEs;
    \item We demonstrated that learned generative mechanisms associated with pathologies are consistent with evidence from neurobiological studies.
\end{enumerate}

\section{Background}
In this section, we introduce the relevant materials on variational auto-encoders, disentangled factorization and the role of supervision. 

\vspace{-10pt}
\subsubsection{Variational auto-encoders}
Variational auto-encoders (VAEs) \cite{Kingma2014AutoEncodingVB} learn a model distribution $p_{\theta}(\mathbf{x}, \mathbf{z})$ that describes the ground-truth data generation process $p(\mathbf{x}, \mathbf{z})$ as first sampling random variables $\mathbf{z}$ from a prior distribution $p(\mathbf{z})$. Then, an observation $\mathbf{x}$ is inferred based on generative factors $p_{\theta}(\mathbf{x} | \mathbf{z})$ yielding
\begin{equation} 
\label{eq:vae_gm}
    p_{\theta}(\mathbf{x}, \mathbf{z}) = p_{\theta}(\mathbf{x} | \mathbf{z}) p(\mathbf{z})
\end{equation}

Here, the conditional distribution is parameterized with neural networks whose learned parameters are denoted with $\theta$. Defining latent variables as jointly independent yields \emph{disentangled factorization} \cite{Locatello2019ChallengingCA} that separates the generative process into human-interpretable \cite{Higgins2017betaVAELB} generative mechanisms. 

\vspace{-10pt}
\subsubsection{Supervised learning}
A label variable $\mathbf{y} \sim p(\mathbf{y})$ can be interpreted as the context that partially governs the generation of an observed variable $\mathbf{x}$. In VAEs, it is reflected by generative factors $p(\mathbf{x}|\mathbf{z}, \mathbf{y})$ of a model. It leads to the joint distribution factorized as follows:

\begin{equation} \label{eq:vae_genmod}
    p_{\theta_1, \theta_2}(\mathbf{x}, \mathbf{z}, \mathbf{y}) = p_{\theta_1}(\mathbf{x}|\mathbf{z}, \mathbf{y}) p_{\theta_2}(\mathbf{z}|\mathbf{y}) p(\mathbf{y})
\end{equation}
where $\theta_1, \theta_2$ are parameters of corresponding model distributions. The equality holds due to the chain rule. Incorporating label information into the model allows learning generative factors corresponding to those labels via $p_{\theta_2}(\mathbf{z}|\mathbf{y})$.

\section{Methods} \label{sec:methods}
\begin{figure*}[ht]
\begin{center}
\centerline{\includegraphics[width=\textwidth]{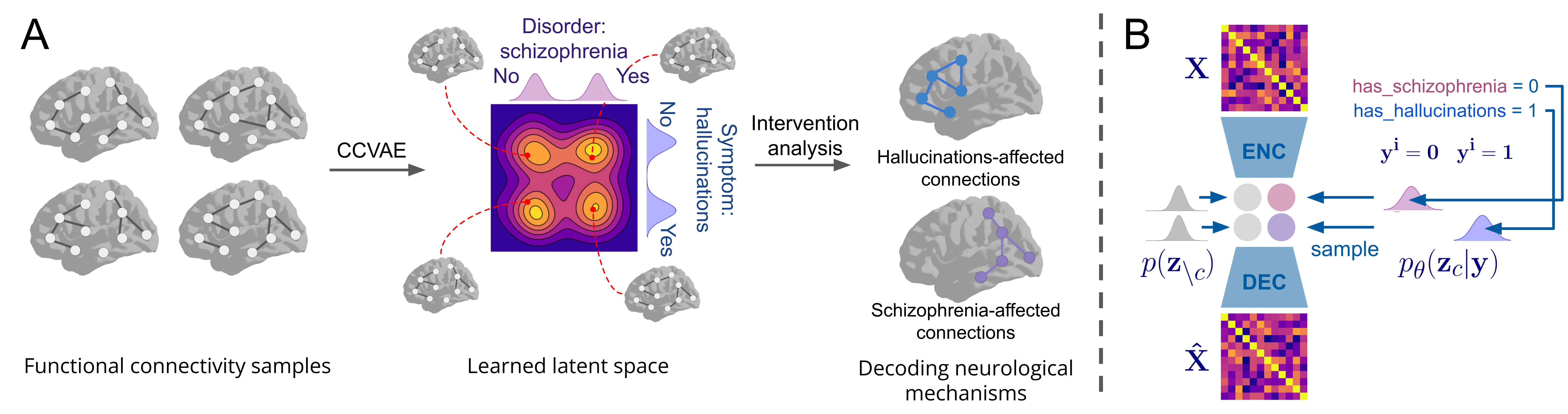}}
\caption{Scheme of the proposed approach (A). We receive EEG data as input and learn a stochastic mapping to the latent space with CCVAEs (B) \cite{Joy2021CapturingLC}. We further manipulate learned generative factors of data to gain insights regarding neurological mechanisms related to the attribute of interest, e.g. a symptom.}
\label{fig:scheme}
\vspace{-25pt}
\end{center}

\end{figure*}

The proposed framework consists of 2 steps (see Fig. \ref{fig:scheme}). First, EEG data is mapped stochastically to the latent space via CCVAEs. The latent space is constructed such that each label is related to a single independent generative factor. Second, we perform an intervention analysis to decode the meaning of label-related generative factors. This way, we get an intuition regarding mechanisms through which labels govern data generation. In our case, we are interested how different pathologies manifest themselves in functional connectivity matrices.

\subsubsection{Characteristic capturing VAEs}
\label{sec:ccvae}

We aim at learning a model of a joint distribution over observed EEG data $\mathbf{x}$, labels (e.g. pathology indicators) $\mathbf{y}$ and latent variables $\mathbf{z}$ partially conditioned to $\mathbf{y}$. Let us assume that $\mathbf{x}$ and $\mathbf{y}$ are conditionally independent given $\mathbf{z}$. Then, the generative model (see Eq. \ref{eq:vae_genmod}) can be rewritten as follows:

\begin{equation*}
    p_{\theta_1, \theta_2}(\mathbf{x},\mathbf{y},\mathbf{z}) = p_{\theta_1}(\mathbf{x}|\mathbf{z}) \: p_{\theta_2}(\mathbf{z} | \mathbf{y} ) \: p(\mathbf{y})
\end{equation*}

We further partition the latent space $\mathbf{z}$ such that one partition $\mathbf{z}_c$ encapsulates label associated characteristics, and the second partition $\mathbf{z}_{\backslash c}$ accounts for shared features of data (as in the vanilla VAEs):
\begin{equation*}
    p_{\theta_2}(\mathbf{z} | \mathbf{y} ) = p_{\theta_2}(\mathbf{z}_c | \mathbf{y} ) \cdot p(\mathbf{z}_{\backslash c})
\end{equation*}

The characteristic partition $\mathbf{z}_c$ is further partitioned so that each label can access only a single latent variable. It guarantees the disentanglement of label information in latent representations. The intractable distribution $p(\mathbf{z} | \mathbf{x},\mathbf{y})$ is conditioned to both observation and label variables. It is approximated with the following inference model:

\begin{equation*}  \label{eq:ccvae_inf_model}
    q_{\phi_1, \phi_2}(\mathbf{z}|\mathbf{x},\mathbf{y}) = \frac{q_{\phi_1}(\mathbf{y}|\mathbf{z}_c) \: q_{\phi_2}(\mathbf{z}|\mathbf{x})}{q_{\phi_1, \phi_2}(\mathbf{y}|\mathbf{x})}
\end{equation*}
where $\phi_1, \phi_2$ are parameters of model distributions. The conditional distribution
\begin{equation*}
   q_{\phi_1, \phi_2}(\mathbf{y} | \mathbf{x}) = \int q_{\phi_1}(\mathbf{y} | \mathbf{z}_c) \: q_{\phi_2}(\mathbf{z}|\mathbf{x}) \: d\mathbf{z} 
\end{equation*}
reflects that observation variables $\mathbf{x}$ and label variables $\mathbf{y}$ are connected via the characteristic partition $\mathbf{z}_c$. Label-related information is captured in an observation $\mathbf{x}$ by the inference model $q_{\phi_2}(\mathbf{z}|\mathbf{x})$. At the same time, classifier $q_\phi(\mathbf{y}|\mathbf{z}_c)$ forces the label-related latent variables $\mathbf{z}_c$ to capture characteristics of those labels.

As for the vanilla VAEs, the model is optimized by maximizing the evidence lower bound \cite{Kingma2014AutoEncodingVB}. In the case of CCVAEs, it is equivalent to maximizing the following objective (see Appendix B.1 of \cite{Joy2021CapturingLC} for derivation):

\begin{equation} \label{eq:ccvae_obj}
    \mathcal{L}(\mathbf{x},\mathbf{y}) = \EX_{q_{\phi_2}(\mathbf{z}|\mathbf{x})}\Bigg[ \frac{q_{\phi_1}(\mathbf{y} | \mathbf{z}_c)}{q_{\phi_1, \phi_2}(\mathbf{y} | \mathbf{x})} log \frac{p_{\theta_1}(\mathbf{x}|\mathbf{z}) \: p_{\theta_2}(\mathbf{z} | \mathbf{y} )}{q_{\phi_1}(\mathbf{y} | \mathbf{z}_c) \: q_{\phi_2}(\mathbf{z}|\mathbf{x})}\Bigg]
    + log \: q_{\phi_1, \phi_2}(\mathbf{y} | \mathbf{x})
\end{equation}
The classification term $log \: q_{\phi_1, \phi_2}(\mathbf{y}|\mathbf{x})$ is essentially a learnable mapping from input data $\mathbf{x}$ to labels $\mathbf{y}$ that goes through the characteristic partition of the latent space $\mathbf{z}_c$. It applies pressure onto the partition to learn label-related characteristics from data and simultaneously performs data classification.

\subsubsection{Intervention analysis}
\label{sec:ia}

The learned generative model forms the bridge between observations $\mathbf{x}$ and their labels $\mathbf{y}$ via latent variables $\mathbf{z}_c$. It allows one to analyze generative factors $p_{\theta_1, \theta_2}(\mathbf{x}|\mathbf{z}, \mathbf{y})$ of data related to those labels. One can explore the relation via intervention analysis. The algorithm for a single \textit{binary} label of interest $\mathbf{y}^i$ is as follows.
First, one fixes every dimension of the latent space $\mathbf{z}$ except the one $\mathbf{z}_c^i$ that corresponds to the label $\mathbf{y}^i$. Next, the value of $\mathbf{z}_c^i$ is sampled from $p_{\theta_2}(\mathbf{z}_c^i | \mathbf{y}^i)$ for each value of $\mathbf{y}^i \in \{0, 1\}$ . As a result, one receives two latent representations $\mathbf{z}_0$, $\mathbf{z}_1$ that vary only in a single dimension $\mathbf{z}_c^i$. Those representations are then reconstructed to the observation space $\mathbf{x}_0 \sim p_{\theta_1}(\mathbf{x}|\mathbf{z} = \mathbf{z}_0)$, $\mathbf{x}_1 \sim p_{\theta_1}(\mathbf{x}|\mathbf{z} = \mathbf{z}_1)$. The procedure is repeated for $N$ times. As a result, one gets multiple pairs of reconstructions $(\mathbf{x}_0, \mathbf{x}_1)$ that are different only to the varied generative factor $\mathbf{z}_c^i$. One further calculates the average difference $\frac{1}{N} \sum_{k=1}^N(\mathbf{x}^k_1 - \mathbf{x}^k_0)$ for each pair, and thus observes how the label $\mathbf{y}^i$ manifests itself in data. 

\section{Related Works} \label{sec:baselines}

The fusion of generative and discriminative models with application to neuroimaging data is an active area of research. \cite{Krishna2020ConstrainedVA} demonstrate that using learned representations leads to more robust classification performance compared to feed-forward neural networks. \cite{Li2020LatentFD} introduce VAEs into feature extraction from multichannel EEG data yielding better accuracy than traditional unsupervised approaches. \cite{Chen2020SemisupervisedDL} use stacked VAEs for semi-supervised learning on EEG data. However, the label information is usually encapsulated by multiple latent variables simultaneously. In this case, label characteristics are smeared across the latent space, thus complicating the analysis of label-related generative factors. It, in turn, limits both the interpretability and explainability of these models. One has to decode and interpret each latent variable and then infer the relation with label variables which is not a trivial task.

Two flavours of VAEs that are commonly applied to EEG data are conditional VAEs \cite{Chen2020SemisupervisedDL}\footnote{Technically, \cite{Chen2020SemisupervisedDL} use stacked VAEs that have two connected latent spaces. One of the spaces is connected to label variables. However, the framework can be seen as an instance of conditional VAEs with a non-trivial structure of the latent space.} and VAEs with downstream classification \cite{Li2020LatentFD}. In both approaches, the latent space is not partitioned with respect to label variables. Hence, compared to CCVAEs, their general disadvantage is reduced interpretability of classification as it is difficult to build a bridge between labels and generative factors. 

\subsubsection{Conditional VAEs}
Conditional VAEs have a graphical model similar to the one of CCVAEs. 
The only difference is that the latent space is not partitioned to labels, i.e. $\mathbf{z} = \mathbf{z_c}$. Learnable parameters are optimized via maximizing the objective Eq. (\ref{eq:ccvae_obj}). The framework allows conditional sampling, so one can use intervention analysis to decode the meaning of learned generative factors. Nevertheless, the interpretation is complicated as a single label variable is connected to each dimension of the latent space.

\subsubsection{VAEs + downstream classification}
The model approximates the joint distribution of observed data and latent variables that is factorized as Eq. (\ref{eq:vae_gm}). The relation between latent variables and labels is built via classifying a latent representation. The model is trained via optimizing the following objective \cite{Krishna2020ConstrainedVA}:
\begin{equation} \label{eq:vae_obj}
\mathcal{L}(\mathbf{x},\mathbf{y}) = \EX_{q_{\phi}(\mathbf{z}|\mathbf{x})}\bigg[ log \: p_\theta(x|z) - D_{KL}(q_\phi(z|x) || p(z)) - BCE(f_\xi(z), y)\bigg]
\end{equation}
where $f_\xi : Z \rightarrow Y$ is a learnable classifier with parameters $\xi$, $BCE$ is binary cross-entropy function. Here, the information about label variables is incorporated into the latent space via pressure applied by a downstream classification task. The model can be seen as a feed-forward deep neural network with additional regularization imposed by the decoder part of VAEs.

\section{Experimental details}
\subsubsection{Experimental study}
The study comprised 29 patients suffering from schizophrenia and 52 healthy controls. 14 subjects out of those 29 indicated the emergence of auditory verbal hallucinations (AVH), i.e. hearing voices with no external stimuli presented. Every participant was right-handed. Six different syllables were spoken to each participant (/ba/, /da/, /ka/, /ga/, /pa/, /ta/) for 500 ms simultaneously to each ear after 200 ms silence period. Meanwhile, the EEG recording was conducted with 64 electrodes where 4 EOG channels were used to monitor eye movements. For each subject, we repeated the procedure multiple times (number of trials for AVH: $68.23 \pm 19.43$; SZ: $68.76 \pm 14.79$ and HC: $71.19 \pm 12.93$). At the preprocessing step, the data was filtered from 20 to 120 Hz according to a protocol described in \cite{Steinmann2017AuditoryVH}. Therefore, only gamma-band frequencies are preserved. Afterwards, all channels were re-referenced to the common average. At last, muscle and visual artefacts were identified and removed. For our experiments, we utilized two parts of a recording: the resting one (first 200 ms with no syllable given) and the listening one (initial 200 ms when syllables were presented). The study of \cite{Steinmann2017AuditoryVH} contains detailed data acquisition and preprocessing information.

\subsubsection{Experimental data}
For each EEG recording $[\zeta_1, \zeta_2, ..., \zeta_{61}] $, we assessed functional connectivity by calculating a correlation matrix:
\begin{equation*}
    \mathbf{x}_{ij} = \frac{ cov(\zeta_i, \zeta_j) } { \sqrt{ var(\zeta_i) \cdot var(\zeta_j) } }
\end{equation*} As a result, functional connectivity matrices play the role of observed data $\mathbf{x}$. We introduce 3 binary labels such that $\mathbf{y} = $ $[\textit{listening},$ $\textit{schizophrenia}$, $\textit{hallucinations}]$. To create a dataset for training models, we use the intra-patient paradigm, i.e. data from the same subject can appear simultaneously in training and test datasets. Thus, data of all the subjects are randomly sampled to form those datasets yielding $9000$ training samples and $2000$ test samples.

\subsubsection{Implementation details}
One can find details regarding the parametrization of distributions in the supplementary material (Section S.1). We release the implementation at \href{https://github.com/maxxxzdn/eegVAE}{GitHub}. For each framework, the parameters $\theta_i, \phi_j$ (and $\xi$ for VAEs) are trained via optimizing the corresponding objective. We use Adam optimizer with a learning rate of $10^{-3}$. The training was performed in mini-batches of size 32 for 100 epochs. All models are trained on an NVIDIA Tesla V100 GPU from the Hemera HPC system of HZDR.

\section{Results \& Discussion}

We found that high-dimensional latent spaces (dim $> 32$) hinder the reproducibility of generative factors learned by CCVAEs. For that reason, we keep the latent space of all models low-dimensional:  $\mathbf{z} \in \mathbb{R}^{5}$ ($\mathbf{z}_{c} \in \mathbb{R}^{3}$, $\mathbf{z}_{\backslash c} \in \mathbb{R}^{2}$ for CCVAEs).

\begin{wraptable}{}{0.55\textwidth}
\vspace{-20pt}
\caption{Comparison of CCVAEs to baseline models in terms of accuracy and disentanglement scores on the test dataset. For each framework, 10 experiments were conducted.}
\centering
\scalebox{0.77}{
\begin{tabular}{cccc}
\hline
Framework & accuracy & SAP score \cite{sap_score} & MIG score \cite{mig_score} \\ \hline
\\[-1em]
CCVAEs & $\mathbf{0.84 \pm 0.01}$ & $\mathbf{0.34 \pm 0.02}$ & $\mathbf{0.04 \pm 0.01}$ \\ 
\\[-1em]
\multicolumn{1}{c}{\begin{tabular}[c]{@{}c@{}}Conditional \\ VAEs\end{tabular}} & $0.69 \pm 0.10$ & $0.07 \pm 0.07$ & $0.01 \pm 0.01$ \\ 
\\[-1em]
\multicolumn{1}{c}{\begin{tabular}[c]{@{}c@{}}VAEs + \\ classification\end{tabular}}  & $0.74 \pm 0.03$ & $0.06 \pm 0.04$ & $0.01 \pm 0.01$\\ \hline
\end{tabular}}
\label{table:baseline}
\vspace{-20pt}
\end{wraptable}

\subsubsection{Results}
As shown in Table \ref{table:baseline}, CCVAEs outperform baseline models in both classification performance and disentanglement (see supplementary material S.2 for details). The framework consistently classifies observed data based on its low-dimensional representation, yielding a low standard deviation of accuracy. Besides, it demonstrates a high level of disentanglement, meaning that each label variable is captured only by a single latent dimension. For CCVAEs, generative factors are disentangled in the latent space by design, leading to the highest score. It is not as high as expected due to the correlation between pathology labels. 

\vspace{-5pt}
\subsubsection{Disentangled latent space}
\label{sec:dls}

\begin{figure}[!tbp]
  \centering
  \begin{minipage}[b]{0.475\textwidth}
    \includegraphics[height=2.5cm, width=1.0\linewidth]{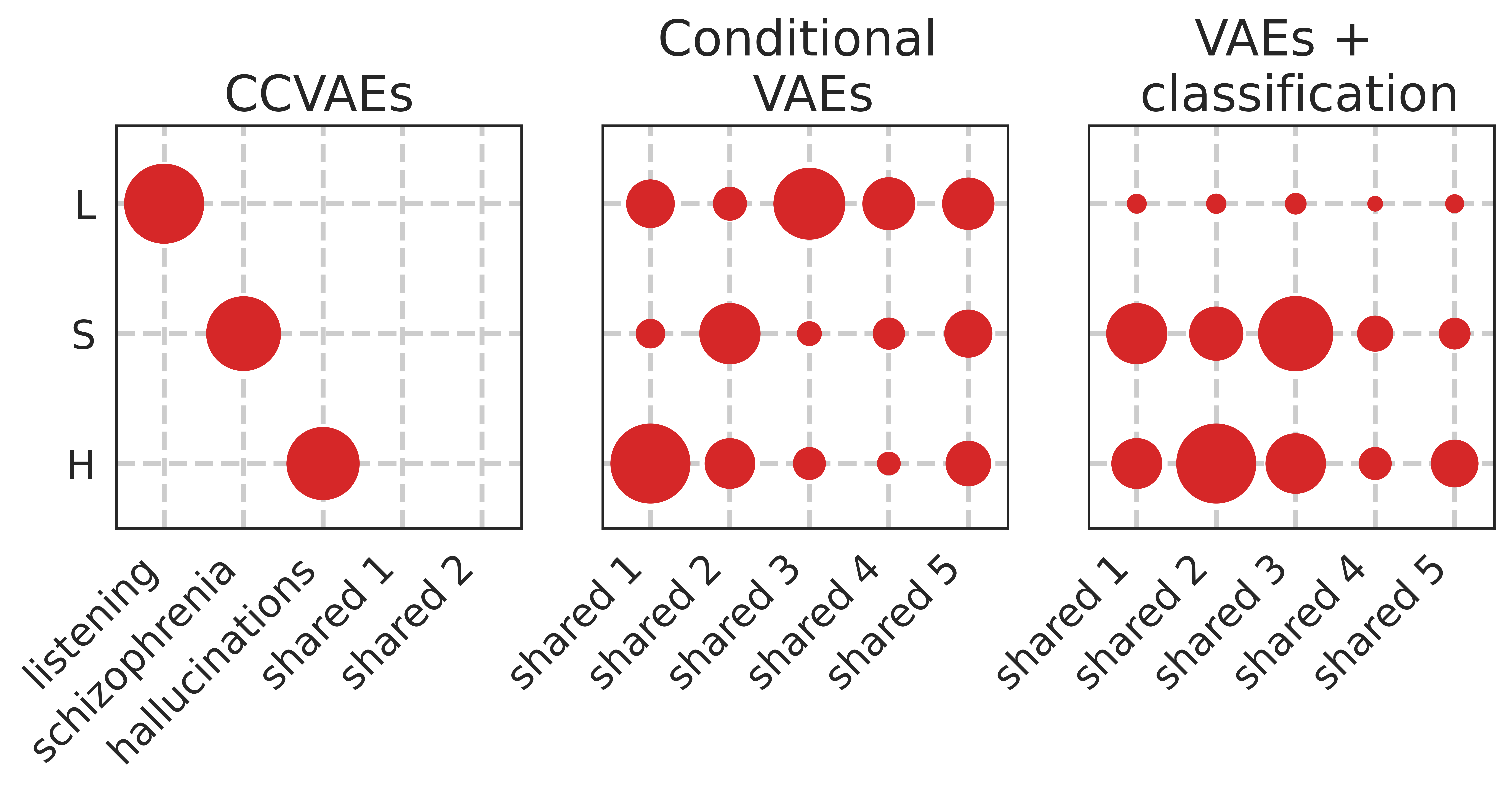}
    \caption{Confusion matrices for different methods. The size of the circle indicates the value of the corresponding element. Rows correspond to label variables (L - \textit{listening}, S - \textit{schizophrenia}, H - \textit{hallucinations}) while columns represent latent generative factors.}
    \label{fig:cm}
  \end{minipage}
  \hfill
  \begin{minipage}[b]{0.475\textwidth}
    \includegraphics[height=2.5cm, width=1.0\linewidth]{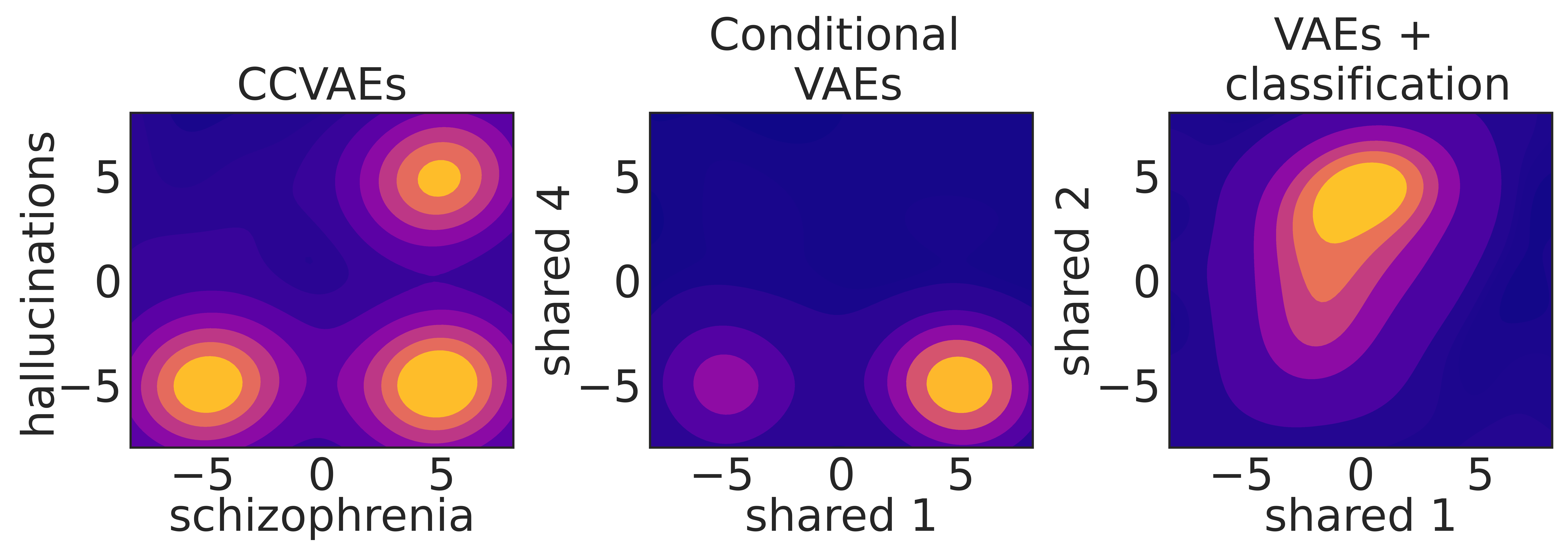}
    \caption{Latent space generated by sampling from the inference model $q_\phi(\mathbf{z}|\mathbf{x})$ of different methods. For CCVAEs, the axes corresponding to pathology variables are shown. For baseline methods, 2 randomly selected dimensions are visualized.}
    \label{fig:ls}
  \end{minipage}
\vspace{-10pt}
\end{figure}

To demonstrate how hard-wired disentanglement affects the latent space learned by CCVAEs, we construct confusion matrices in the following way. We compute latent representation for each data point in the test dataset and intervene (i.e. randomly change its value) upon a single dimension. We further observe how log-probabilities assigned by a pre-trained classifier change due to the intervention for each label. We calculate the difference for each label-latent pair yielding a confusion matrix (see Fig. \ref{fig:cm}). The optimal result would be one non-zero element per row (i.e. label), which means that each label corresponds to only a single generative factor. This is the case of CCVAEs, where one can observe one-to-one dependence between label variables and corresponding generative factors. This leads to latent representations being robust to variations in data generative factors, as those are independent by design. At the same time, the characteristics of labels are entangled within latent spaces of baseline frameworks. Hence, it is difficult to disentangle the influence of a label from other generative factors, which severely hinders interpretability.

\vspace{-5pt}
\subsubsection{Posterior distribution}
\label{sec:psa}

We further compare latent spaces learned by each framework. To visualize the latent space, we sample  $\mathbf{z} \sim q_\phi(\mathbf{z}|\mathbf{x})$ for multiple $\mathbf{x}$ from the test dataset for each model. The result is shown in Fig. \ref{fig:ls}. In the case of CCVAEs, the distribution has three modes corresponding to subject cohorts in data (healthy, schizophrenia, schizophrenia followed by AVH). The separation is caused by the influence of the conditional prior and the classifier, aiming to separate representations encoding different label combinations. In the case of baseline frameworks, there is no strict regularisation that preserves label information within a partition of the latent space. As a result, features of data encoded by their latents are shared between cohorts of subjects (thus one or two modes). We discovered that both baseline methods often fail to jointly learn a low-dimensional representation and classify labels when the pressure on the KL divergence term in the loss objective is high. The problem is partially solved by introducing a scaling factor $\beta$ for the term \cite{Higgins2017betaVAELB}. However, reducing the pressure might lead to untrustable reconstruction if prior $p(\mathbf{z})$ is not sufficiently close to the inference model $q(\mathbf{z}|\mathbf{x})$. This is not the case for CCVAEs that do not require any manual fine-tuning and operate stably with low-dimensional latent spaces. 
\subsubsection{Analyzing pathological mechanisms}
\label{sec:avh}

\begin{wrapfigure}{}{0.5\textwidth}
\vspace{-20pt}
\centering
\begin{subfigure}{0.24\textwidth}
    \includegraphics[width=2.5cm, height = 2.5cm]{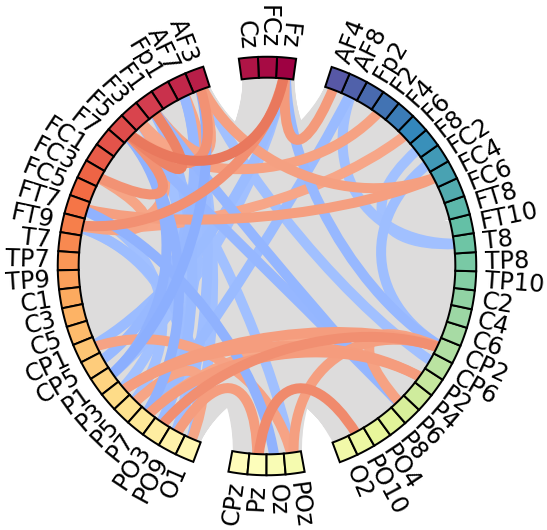}
    \caption*{$\mathbf{y}^i = $ schizophrenia}
\end{subfigure}
\begin{subfigure}{0.24\textwidth}
    \includegraphics[width=2.9cm, height = 2.5cm]{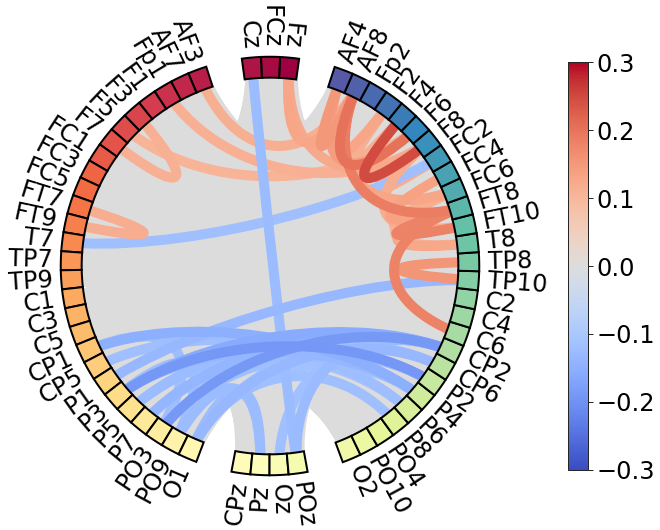}
    \caption*{$\mathbf{y}^i = $ hallucinations}
\end{subfigure}
\caption{Average difference in reconstructions of functional connectivity matrices when intervening on a single label: \textit{schizophrenia} (Left) and \textit{hallucinations} (Right). Connections that are stronger when a disorder is presented are shown in red; otherwise, blue. For clarity, we visualize only 40 connections with the highest absolute value.} \label{fig:var_con}
\vspace{-20pt}
\end{wrapfigure}

We further investigate what connections are affected when intervening upon a single label dimension via intervention analysis (Fig. \ref{fig:var_con}, see supplementary material S.3 for computation details). The model associates the emergence of AVH with alterations in frontotemporal brain areas (the highest positive difference), which have been repeatedly observed in prior studies \cite{Jardri2011CorticalAD}, \cite{Lavigne2015LeftdominantTH}. The salient connections are mainly located in the right hemisphere, which is supported by the fMRI study of \cite{Hwang2021AuditoryHA}. The model also points toward reduced connectivity between hemispheres. It is coherent with the current hypothesis (see \cite{Steinmann2017AuditoryVH} for review) that connects the emergence of auditory verbal hallucinations with the interhemispheric miscommunication during auditory processing. Overall, the model can at least partially reconstruct the neurological mechanism of the symptom for functional connectivity. To explain the emergence of schizophrenia, the model focuses mainly on the left hemisphere. It is not surprising since the auditory function is left-lateralized for right-handed people \cite{Papathanassiou2000ACL}, \cite{Flinker2015RedefiningTR}. It would be an interesting direction for further studies to apply CCVAEs to learn the mechanisms of particular symptoms of the composite disorder (e.g. hallucinations, delusions, etc.).

\vspace{-5pt}
\section{Conclusion}
We demonstrated how to apply the framework of characteristic capturing variational auto-encoders to EEG data analysis. The method encapsulates and disentangles the characteristics associated with different pathologies in the latent space. As generative factors are independent by design, one can decode their meaning and discover how those pathologies alter observed data. It leads to improved interpretability coupled with the high classification performance of neural networks. The framework is not limited to functional connectivity analysis or EEG data and can be easily adapted to different neuroimaging modalities.

%
%
%
%

\newpage
\renewcommand{\thesection}{S}

\section*{Supplementary material}
\subsection{Parametrization of model distributions}

\begin{wrapfigure}{}{0.4\textwidth}
\vspace{-20pt}
\centering
\includegraphics[width=.8\linewidth]{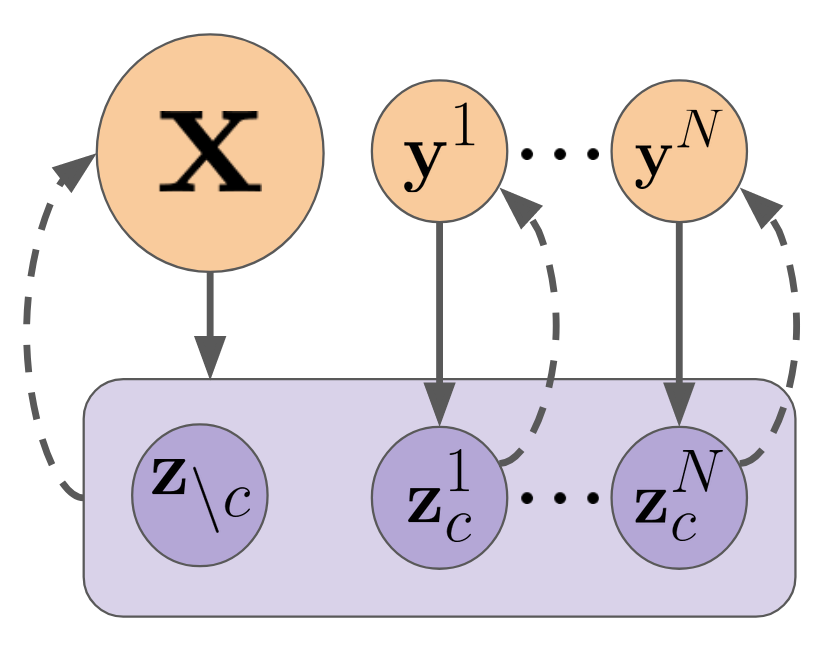}
\caption{Graphical model of CCVAEs. Generative models are denoted with dashed lines, whilst solid lines represent inference models. Observed variables are depicted in orange, and latent variables are shown in purple.}
\label{fig:gm_ccvae}
\vspace{-15pt}
\end{wrapfigure}

The graphical model of CCVAEs is shown in Fig. \ref{fig:gm_ccvae}. We define the generative and inference models as follows. Every conditional distribution is represented as a Normal distribution except except $q_{\phi_1}(\mathbf{y} | \mathbf{z})$ which is defined as a Binomial distribution. Parameters of each distribution are defined as deterministic functions of conditional variables. We represent functional connectivity matrices $\mathbf{x}$ as images of size $1 \times 64 \times 64$ by adding zero padding to initial $61 \times 61$ matrices. Inference model $q_{\phi_2}(\mathbf{z}|\mathbf{x})$ and generative model $p_{\theta_1}(\mathbf{x}|\mathbf{z})$ are parameterized with CNNs that approximate $\mu(\circ), \sigma(\circ)$, where $\circ$ denote corresponding conditional variables. In the case of CCVAEs, parameters of the conditional prior $p_{\theta_2}(\mathbf{z}_c^i | \mathbf{y}^i)$ are implemented as linear element-wise transformations: $\mu_c^i = w^i \cdot \mathbf{y}^i + b^i$, $\sigma_c^i = 1$. For conditional VAEs, those parameters are represented by an MLP. The classifier $q_{\phi_1}(\mathbf{y} | \mathbf{z_c})$ is parameterized with sigmoid for CCVAEs, and an MLP for conditional VAEs. The classification function $f_\xi$ is implemented as an MLP. Architectural choices for dim $ \mathbf{z} = 5$, dim $ \mathbf{y} = 3$ are given in Table 1.

\begin{table}[hptb]
\caption{Architecture of models.}
\small
\centering
\scalebox{0.9}{
\begin{tabular}{cc}
\hline
Encoder $q_{\phi_2}(\mathbf{z}|\mathbf{x})$         & Decoder $p_{\theta_1}(\mathbf{x}|\mathbf{z})$  \\ \hline
Input $64 \times 64 \times 1$ channel image & Input $ \in \mathbb{R}^5$ \\ 
Conv2D $8 \times 1 \times 5 \times 5$ (stride $2$) \& ReLU    & Linear $5 \times 128$ \& ReLU \\ 
Conv2D $16 \times 8 \times 5 \times 5$ (stride $2$) \& ReLU   & ConvT2D $8 \times 16 \times 5 \times 5$ (stride $2$) \& ReLU  \\ 
Conv2D $16 \times 16 \times 5 \times 5$ (stride $2$) \& ReLU   & ConvT2D $16 \times 16 \times 5 \times 5$ (stride $2$) \& ReLU \\ 
Conv2D $16 \times 8 \times 5 \times 5$ (stride $2$) \& ReLU    & ConvT2D $16 \times 8 \times 5 \times 5$ (stride $2$) \& ReLU \\ 
$2 \times$ Linear $128 \times 5$ & ConvT2D $8 \times 1 \times 5 \times 5$ (stride $2$) \\
\hline
\end{tabular}}
\bigbreak
\begin{tabular}{cc}
\hline
Classifier $q_{\phi_1}(\mathbf{y}|\mathbf{z}_c)$         & Conditional Prior $p_{\theta_2}(\mathbf{z_c}|\mathbf{y})$  \\ \hline
Input $ \in \mathbb{R}^3$ & Input $ \in \mathbb{R}^3$ \\ 
Sigmoid   & $3 \times 3$ Element-wise \\ 
\hline
\end{tabular}
\end{table}

\subsection{Disentanglement metrics}

We compare different methods in terms of disentanglement, i.e. we measure how sensitive a single latent variable is to changes in generative factors \cite{Higgins2017betaVAELB}. A high disentanglement score is achieved when there is a one-to-one correspondence between latent variables and generative factors. We use two common scores: SAP score \cite{sap_score} and MIG score \cite{sap_score}. Both metrics do not demand having full control over the data generative process and are relatively easy to compute.

The SAP score is computed in two steps. First, one calculates the classification score of predicting $j$’th factor using only $i$’th latent, yielding a score matrix. After that, the difference between the top two entries for each column is evaluated and averaged. To calculate the MIG score, one first computes a matrix where each element is the empirical mutual information between a latent variable and a ground truth generative factor. Next, as for the SAP score, one computes the column-wise mean of the difference between the top two latent variables with the highest mutual information.

\subsection{Intervention analysis}

In Section 6, we investigate what connections are affected when intervening upon a single label dimension. We utilize the intervention analysis, algorithm of which we described in Section 3. We are particularly interested in studying 1) how schizophrenia alters functional connectivity compared to healthy controls during auditory processing; 2) how having auditory verbal hallucinations manifest themselves in the reaction of a person with schizophrenia to an external sound. The first task is achieved by keeping \textit{hallucinations} $= 0$, \textit{listening} $= 1$ fixed and varying \textit{schizophrenia}. The second one is tackled by fixing \textit{schizophrenia} $= 1$, \textit{listening} $= 1$ and varying \textit{hallucinations}. For each scenario, we generate $N = 1000$ pairs of samples. Those pairs are different only with respect to a single latent variable encoding the characteristic of a label of interest. The difference between those pairs for each connection is further averaged.

\end{document}